\documentclass{article}
\usepackage{ijcai18}

\usepackage{times}
\usepackage{soul}
\usepackage{url}
\usepackage[hidelinks]{hyperref}
\usepackage[utf8]{inputenc}
\usepackage[small]{caption}

\usepackage{floatrow}
\usepackage{graphicx}

\title{Explanatory Masks for Neural Network Interpretability}

\author{Lawrence Phillips\textsuperscript{1}, Garrett Goh\textsuperscript{2}, Nathan Hodas\textsuperscript{1}, \\
\\
\textsuperscript{1} Computing \& Analytics Division, Pacific Northwest National Lab \\
\textsuperscript{2} Advanced Computing, Mathematics \& Data Division, Pacific Northwest National Lab \\
\texttt{\{lawrence.phillips, garret.goh, nathan.hodas\}@pnnl.gov}
}

\begin{document}

\maketitle

\begin{abstract}
  Neural network interpretability is a vital component for applications across a wide variety of domains. In such cases it is often useful to analyze a network which has already been trained for its specific purpose. 
In this work, we develop a method to produce explanation masks for pre-trained networks. The mask localizes the most important aspects of each input for prediction of the original network. Masks are created by a secondary network whose goal is to create as small an explanation as possible while still preserving the predictive accuracy of the original network. We demonstrate the applicability of our method for image classification with CNNs, sentiment analysis with RNNs, and chemical property prediction with mixed CNN/RNN architectures. 
\end{abstract}

\section{Introduction}

Network interpretability remains a required feature for machine learning systems in many domains. A great deal of recent work has been done to try to shed light on deep learning models, producing methods that create local explanations~\cite{ribeiro2016should}, following gradients~\cite{selvaraju2016grad}, investigating input perturbations~\cite{fong2017interpretable}, and even generate textual explanations~\cite{hendricks2016generating}. Many of these techniques involve creating an input \emph{mask}, assigning weights to individual inputs such that a weighted visualization of the input can be created.

In this work, we explore an alternate method for generating such an explanatory mask. Our goal is to generate an explanatory input mask with two properties: 1) the explanation should be as minimal as possible, 2) the explanation should highlight the portions of the input most necessary to ensure predictive accuracy. To keep the explanation mask as small as possible we make use of a variety of regularization techniques. To keep predictive accuracy we force the mask to generate accurate outputs when passed through the original network.

\section{Explanation Masks}
\begin{figure*}[!htbp]
\centering
\includegraphics[width=0.7\textwidth]{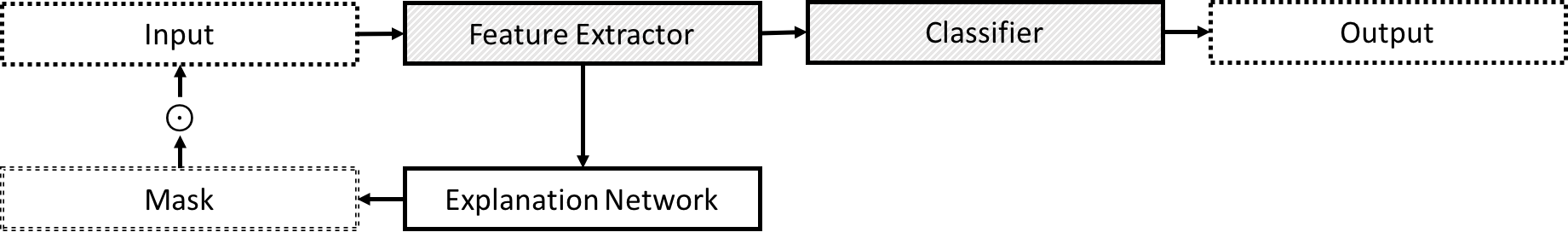}
\caption{\small Explanation mask architecture. The pre-trained network on top is divided into a feature extraction and classification component. The output of the feature extractor feeds into an explanation network, producing a mask that is multiplied element-wise with the input. The masked input then is fed through the original network to produce a final output. Shading indicates frozen weights.}
\label{fig:mask}
\end{figure*}



Given a pre-trained network, the goal of the explanation mask is to identify which inputs are most necessary for creating accurate decisions. 
To do this, we make use of a secondary, \emph{explanation} network, which learns to generate explanation masks in an end-to-end fashion. By training the network, we are able to ensure that the generated mask does indeed carry predictive accuracy. Further, by implementing regularization over the mask, we ensure that the mask is as minimal as possible while still being accurate. This trade-off between accuracy and the size of the mask is one which can be explored on a per-case basis by changing the amount of regularization imposed.


The structure of the explanation network can be of any form fit to the problem task which leaves the method quite flexible. A schematic of the overall system is given in Figure~\ref{fig:mask}. First, the pre-trained network is split into a feature extraction and classification component. This can be done arbitrarily, but can often be done after a feature \emph{bottleneck} in the original network. The output of this feature extraction step is then fed into the explanation network. Given the features describing the input, the explanation network can then generate a mask of the same size as the input. The mask should have weights between 0 and 1 and is multiplied element-wise against the original input.

Once the input has been masked, it can then be fed through the entirety of the original network, both feature extraction and classification components, to produce a final output. To train the explanation network, we freeze the weights of the original network. This ensures that the explanation mask generated is an accurate reflection of what inputs are necessary for the pre-trained network. Training can be accomplished end-to-end through the use of backpropagation over the non-frozen weights.

\subsection{Related Work}

Over the past few years there have been many attempts to shed light on the inner workings of neural networks. One of the most general approaches has been the introduction of attention mechanisms as initially popularized in machine translation \cite{bahdanau2015neural}. Attention mechanisms operate similarly to our explanation masks in that they result in a mask where every element in the input is weighted between 0 and 1 and combined through a weighted sum or product. Traditional attention is trained as a single network which benefits from the end-to-end nature of deep learning but also requires attention to be added initially. This means that while attention mechanisms are often worth incorporating when creating a model, they cannot be used to explain a pre-trained model's decisions if the original model did not have an attention mechanism.

An intriguing alternative to basic attention is the ability to generate natural language explanations to account for a model's decisions. This idea is explored in Hendricks et al. \shortcite{hendricks2016generating}, where an image classifier was trained along with a sentence generator in order to create a short, explanatory sentence along with every classification. A limitation of this work is the weakly-enforced correspondence between image and explanation, which can result in generic explanations that do not reference portions of the image. This is refined in Hendricks et al. \shortcite{hendricks2017grounding} where explanations are more carefully trained to be tailored to a specific image. While powerful, these models require complicated training through reinforcement learning and are only applicable for classification datasets which contain ground truth explanations for each class.

A last group of methods attempt to explain classifier predictions through perturbing data and modeling how this affects the base model. This is explored in Fong et al. \shortcite{fong2017interpretable} where images are perturbed through effects such as blurring in a systematic fashion to reduce the model's classification accuracy. In this way, the authors are able to identify regions of the image which are important to classification. A similar method was introduced in Selvaraju et al. \shortcite{selvaraju2016grad} by exploring model gradients. Lastly, LIME \cite{ribeiro2016should} has been a popular approach for creating a ``interpretable" models, such as sparse linear models, that are locally consistent with more complex black-box models. The downside of this method is that it necessarily removes information from the original model and some have argued that simple models are less interpretable than one might original assume \cite{lipton2016mythos}.

\section{Experiments}

\subsection{Image Classification}

We make use of the CIFAR10 image classification dataset~\cite{krizhevsky2009learning} to test our models ability on visual data. CIFAR10 is made up of 60,000 images from 10 classes, with 50,000 used for training and 10,000 as a test set. As a base model, we use ResNet164 v2~\cite{he2016identity} which is a 164-layer residual network. After training, the model achieves a test set accuracy of 94.04\%.

The ResNet architecture generates a vector of length 256 to represent the input image before making a final dense softmax decision regarding its category. We take this bottleneck vector as the input to the explanation network. The 256-dimension vector is reshaped into a $16 \times 16$ matrix and passed through two convolutional blocks. Each block is made up of 4 2D convolutional blocks with \emph{tanh} activations. Padding ensures the size of the image remains constant and a filter size of 64 is used. A residual connection~\cite{skipconnections} is used where the first and final convolutional layers have their activations summed. After the first block, the mask is upsampled to $32 \time 32$, the size of the original image. After the second block, the explanation network ends with a single 2D convolution with filter size 1 and sigmoid activation.

\begin{figure*}[!htbp]
\centering
\includegraphics[width=0.6\textwidth]{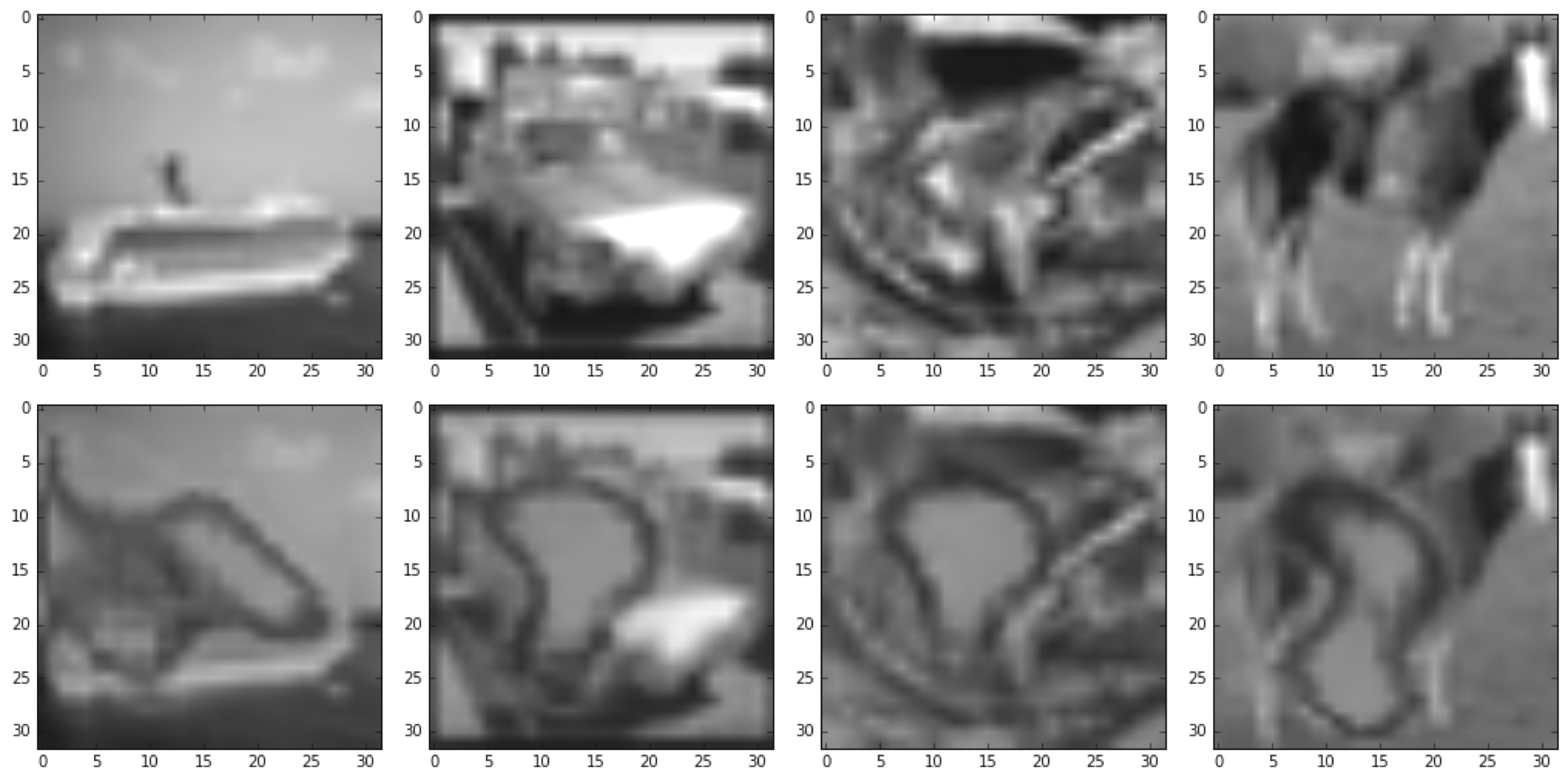}
\caption{\small CIFAR10 images (top) along with their learned explanation masks (bottom).}
\label{fig:cifar}
\end{figure*}

As described above, the mask is multiplied by the original input and is applied equally across all 3 color channels. Sparsity in the mask is encouraged through regularization with l2 = 1e-4. We present original CIFAR10 images along with their learned masks in Figure~\ref{fig:cifar}

\subsection{Sentiment Analysis}

To assess how well the explanation masks work on natural language tasks, we look at the IMDB review sentiment task~\cite{imdb}. This involves 50,000 movie reviews rated as either positive or negative. We preprocess the input so that it has a vocabulary size of 20,000 and all infrequent words are replaced with a special OOV token. We further reduce the cap the maximum length of each review at 500 words. Our base model uses 100-dimensional, pre-trained GloVe embeddings with 40\% word-level dropout to prevent overfitting. This is followed by a bidirectional GRU layer of width 256. The output of this layer is averaged over timesteps and passed to a 128-dimension fully connected layer with relu activation and a final dense softmax performs the final classification. We train the network to optimize binary crossentropy with the Adam optimizer using the standard 50/50 train/test split. After training for 10 epochs we achieve a test set accuracy of 84.7\%.

\begin{figure*}[h]
\includegraphics[width=0.9\textwidth]{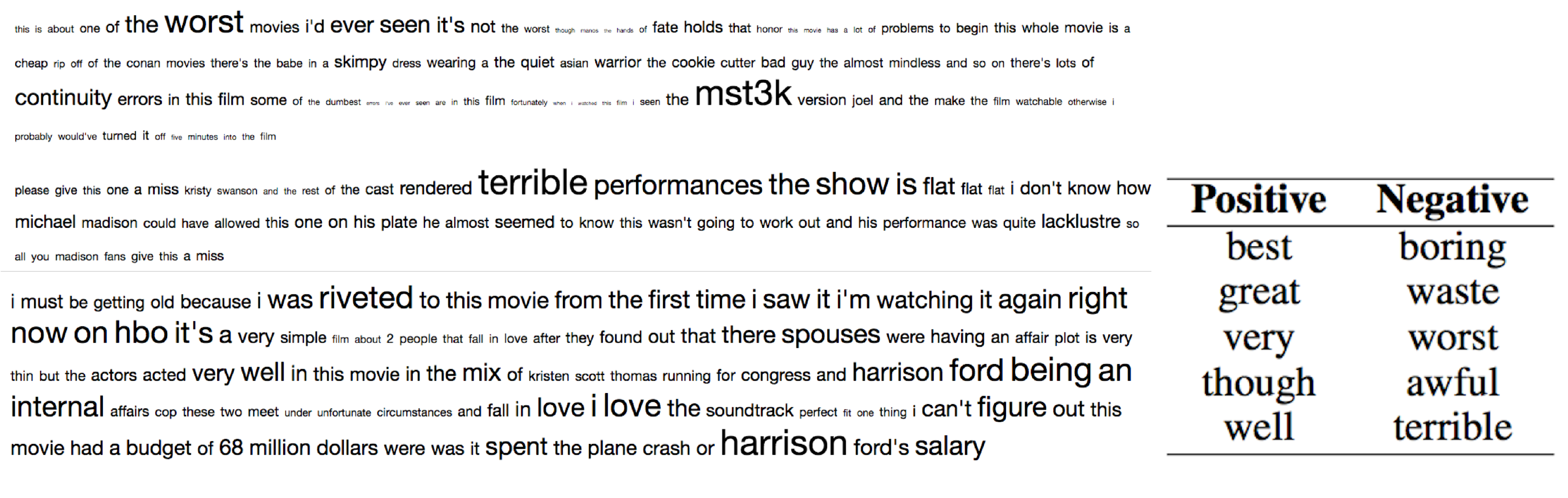}
\caption{\label{fig:imdb}\small Left: IMDB reviews with words scaled in height based on explanation mask weights. Right: Top 5 words of positive and negative sentiment.}
\end{figure*}

For our explanation network we use the full output of the bidirectional GRU layer as input. The explanation mask is calculated by using a 2-layer bidirectional GRU of width 100. Each timestep's output is passed to a fully connected layer of size 256 with relu activation before a linear layer outputs a single value for the timestep, representing the weight of that word in the mask. The mask is regularized by penalizing its entropy. After training for 10 epochs we achieve a test set accuracy of 81.8\%. In Figure~\ref{fig:imdb} we present example sentences with the size of each word scaled to the mask weights. We also explore how well the model is able to capture general positive and negative trends by looking at which words receive the most attention in positive and negative reviews, which we display on the right of Figure~\ref{fig:imdb}.

From the example sentences in Figure~\ref{fig:imdb}, one can see that the model is able to attend to words and phrases which indicate both positive and negative sentiment. In the first sentence, the model recognizes that the word {\em mst3k}, referencing the TV show ``Mystery Science Theater 3000", is highly indicative of negative sentiment. In the final example, the name {\em Harrison} is identified as highly positive. Attention is more often placed over phrases, indicating that the model has identified that combinations of words are indicative of sentiment. Again in the final example, we see the phrase ``right now on HBO" given high attention. In the second example, attention spans two sentences over ``rendered terrible performances. The show is flat".

\subsection{Chemical Solubility}
\begin{figure*}[!htbp]
\centering
\includegraphics[scale=1.2]{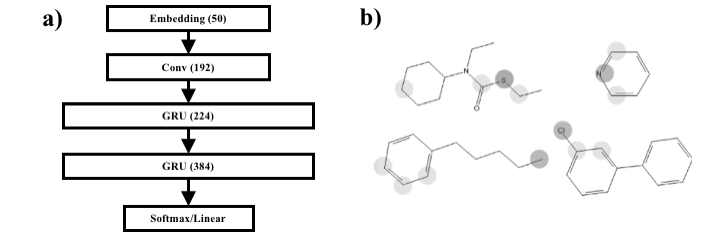}
\caption{\small (a) Illustration of the SMILES2vec CNN-GRU architecture. (b) The explanation mask of SMILES2vec validates established knowledge by focusing on atoms of known hydrophobic and hydrophilic groups, colored circles of increasing darkness indicate increasing attention.}
\label{fig:smiles}
\end{figure*}

As a final proof-of-concept, we explore the ability of our method to work for networks trained on small data. To do this we make use of the SMILES2vec model~\cite{goh2017smiles2vec} to predict chemical solubility on the ESOL dataset~\cite{wu2018moleculenet}. The ESOL dataset is quite small, with only 1,128 data points and differs from the previous tasks in that it involves regression rather than classification.

The SMILES2vec model is a mixed CNN-GRU architecture which takes in chemical compounds as strings represented using SMILES~\cite{weininger1988smiles} and predicts chemical solubility. The network is structured as in Figure~\ref{fig:smiles}a. The network is trained as in~\cite{goh2017smiles2vec} and achieves a test RMSE of 0.846. The explanation network takes as input the embedded SMILES representation and then outputs a character-level attention mask which is applied equally across all embedding dimensions for each character. The network is structured as a 20 layer convolutional residual network with SELU non-linearities. Padding is used for each convolution to ensure the input size remains constant. The final layer of the explanation network is a 1D convolution of length 1 with batch normalization followed by a softplus activation. We regularize the explanation network's output through L1 and L2 loss set to 1e-3 and 1e-4 respectively.  The explanation network is trained until convergence and has a test RMSE of 0.831, slightly better than the original network alone.

To test the model's interpretations, we examined the output qualitative and quantitatively. In Figure~\ref{fig:smiles}b, we highlight the explaination mask, with darker circles indicating more attention from the network. For soluble molecules, the network properly attends to atoms which are part of hydrophilic groups, which increase solubility. The opposite is true of insoluble molecules, as in the bottom right, where the network attends to the Cl molecule as part of a hydrophobic group. To quantify these patterns, we split the data between soluble ($\leq$-1.0) and insoluble ($\geq$-5.0) compounds. The network should attend more to the atoms O and N for soluble compounds and C, F, Cl, Br, and I for insoluble. We identified the top-3 characters attended to in each explanation mask and computed the top-3 accuracy for identifying these particular atoms. The network achieves a top-3 accuracy of 88\%, indicating that the attention mask does reliably identify atoms known to affect solubility.

\section{Conclusion}

Through experiments over three diverse domains and network architectures we have shown how pre-trained networks might be analyzed by means of explanation masks. Although generating explanation masks requires a small amount of training for the explanation network, the method is quite flexible in its structure and can be applied to most pre-trained networks. While there are many possible approaches to model explanation, ours explicitly focuses on the aspects of the input which are most necessary for accurate prediction.


\bibliographystyle{named}
\bibliography{explanations}

\begin{thebibliography}{}

\bibitem[\protect\citeauthoryear{Bahdanau \bgroup \em et al.\egroup
  }{2015}]{bahdanau2015neural}
Dzmitry Bahdanau, Kyunghyun Cho, and Yoshua Bengio.
\newblock Neural machine translation by jointly learning to align and
  translate.
\newblock In {\em ICLR}, 2015.

\bibitem[\protect\citeauthoryear{Fong and
  Vedaldi}{2017}]{fong2017interpretable}
Ruth~C Fong and Andrea Vedaldi.
\newblock Interpretable explanations of black boxes by meaningful perturbation.
\newblock {\em arXiv preprint arXiv:1704.03296}, 2017.

\bibitem[\protect\citeauthoryear{Goh \bgroup \em et al.\egroup
  }{2017}]{goh2017smiles2vec}
Garrett~B Goh, Nathan~O Hodas, Charles Siegel, and Abhinav Vishnu.
\newblock Smiles2vec: An interpretable general-purpose deep neural network for
  predicting chemical properties.
\newblock {\em arXiv preprint arXiv:1712.02034}, 2017.

\bibitem[\protect\citeauthoryear{He \bgroup \em et al.\egroup
  }{2016a}]{skipconnections}
Kaiming He, Xiangyu Zhang, Shaoqing Ren, and Jian Sun.
\newblock Deep residual learning for image recognition.
\newblock In {\em Proceedings of the IEEE conference on computer vision and
  pattern recognition}, pages 770--778, 2016.

\bibitem[\protect\citeauthoryear{He \bgroup \em et al.\egroup
  }{2016b}]{he2016identity}
Kaiming He, Xiangyu Zhang, Shaoqing Ren, and Jian Sun.
\newblock Identity mappings in deep residual networks.
\newblock In {\em European Conference on Computer Vision}, pages 630--645.
  Springer, 2016.

\bibitem[\protect\citeauthoryear{Hendricks \bgroup \em et al.\egroup
  }{2016}]{hendricks2016generating}
Lisa~Anne Hendricks, Zeynep Akata, Marcus Rohrbach, Jeff Donahue, Bernt
  Schiele, and Trevor Darrell.
\newblock Generating visual explanations.
\newblock In {\em European Conference on Computer Vision}, pages 3--19.
  Springer, 2016.

\bibitem[\protect\citeauthoryear{Hendricks \bgroup \em et al.\egroup
  }{2017}]{hendricks2017grounding}
Lisa~Anne Hendricks, Ronghang Hu, Trevor Darrell, and Zeynep Akata.
\newblock Grounding visual explanations.
\newblock In {\em NIPS Interpretable ML Symposium}, 2017.

\bibitem[\protect\citeauthoryear{Krizhevsky and
  Hinton}{2009}]{krizhevsky2009learning}
Alex Krizhevsky and Geoffrey Hinton.
\newblock Learning multiple layers of features from tiny images.
\newblock 2009.

\bibitem[\protect\citeauthoryear{Lipton}{2016}]{lipton2016mythos}
Zachary~C Lipton.
\newblock The mythos of model interpretability.
\newblock {\em arXiv preprint arXiv:1606.03490}, 2016.

\bibitem[\protect\citeauthoryear{Maas \bgroup \em et al.\egroup }{2011}]{imdb}
Andrew~L Maas, Raymond~E Daly, Peter~T Pham, Dan Huang, Andrew~Y Ng, and
  Christopher Potts.
\newblock Learning word vectors for sentiment analysis.
\newblock In {\em Proceedings of the 49th annual meeting of the association for
  computational linguistics: Human language technologies}, volume~1, pages
  142--150, 2011.

\bibitem[\protect\citeauthoryear{Ribeiro \bgroup \em et al.\egroup
  }{2016}]{ribeiro2016should}
Marco~Tulio Ribeiro, Sameer Singh, and Carlos Guestrin.
\newblock Why should i trust you?: Explaining the predictions of any
  classifier.
\newblock In {\em Proceedings of the 22nd ACM SIGKDD International Conference
  on Knowledge Discovery and Data Mining}, pages 1135--1144. ACM, 2016.

\bibitem[\protect\citeauthoryear{Selvaraju \bgroup \em et al.\egroup
  }{2016}]{selvaraju2016grad}
Ramprasaath~R Selvaraju, Michael Cogswell, Abhishek Das, Ramakrishna Vedantam,
  Devi Parikh, and Dhruv Batra.
\newblock Grad-cam: Visual explanations from deep networks via gradient-based
  localization.
\newblock {\em See https://arxiv. org/abs/1610.02391 v3}, 7(8), 2016.

\bibitem[\protect\citeauthoryear{Weininger}{1988}]{weininger1988smiles}
David Weininger.
\newblock Smiles, a chemical language and information system. 1. introduction
  to methodology and encoding rules.
\newblock {\em Journal of chemical information and computer sciences},
  28(1):31--36, 1988.

\bibitem[\protect\citeauthoryear{Wu \bgroup \em et al.\egroup
  }{2018}]{wu2018moleculenet}
Zhenqin Wu, Bharath Ramsundar, Evan~N Feinberg, Joseph Gomes, Caleb Geniesse,
  Aneesh~S Pappu, Karl Leswing, and Vijay Pande.
\newblock Moleculenet: a benchmark for molecular machine learning.
\newblock {\em Chemical Science}, 9(2):513--530, 2018.

\end{thebibliography}

\end{document}